\documentclass[a4paper, 10pt, conference]{ieeeconf}      
\usepackage{FG2026}
\usepackage{graphicx}
\usepackage{amsmath} 
\usepackage{amssymb}
\usepackage{subcaption}
\FGfinalcopy 

\IEEEoverridecommandlockouts                              
\def\FGPaperID{9} 

\title{\LARGE \bf
Age-Defying Face Recognition with Transformer-Enhanced Loss
}

\author{\parbox{16cm}{\centering
    {\large Pritesh Prakash and Anoop Kumar Rai }\\
    {\normalsize
    Central Research Laboratory, Bharat Electronics Limited, Ghaziabad, India, 201010}}
}

\ifFGfinal
\else
\author{Anonymous FG2026 submission\\ Paper ID \FGPaperID \\}
\fi

\begin{document}
\maketitle
\begin{abstract}

Aging poses a significant challenge in face recognition, as changes in skin texture and tone alter facial features over time. This makes it difficult to compare images of the same individual taken years apart, such as in long-term identification scenarios. Transformer networks can effectively preserve sequential spatial relationships affected by aging. This paper introduces a technique that incorporates a transformer network as an additive loss function for face recognition. This research aims to analyze the transformer’s behavior when processing convolution outputs, with CNN feature maps arranged sequentially. These sequential vectors can mitigate the impact of aging-related changes, such as wrinkles or sagging skin. The learned features are more age-invariant, enhancing the discriminative power of standard metric loss embeddings. Using this technique, we combine transformer loss with various metric loss functions to evaluate their combined impact. We observe that this configuration enables the network to achieve state-of-the-art (SoTA) results on age-variant datasets, such as CA-LFW and AgeDB. This research expands the application of transformers in machine vision and paves the way for their innovative use as a loss function in machine learning.
\end{abstract}

\section{Introduction}
\begin{figure}
    \centering
    \fbox{
        \subcaptionbox{\label{p_1}}{
            \includegraphics[scale=0.4]{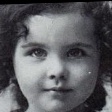}
            \includegraphics[scale=0.4]{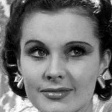}
        }    
        \subcaptionbox{\label{p_2}}{
            \includegraphics[scale=0.4]{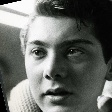}
            \includegraphics[scale=0.4]{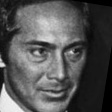}
        }
    }
    \fbox{
        \subcaptionbox{\label{p_3}}{
            \includegraphics[scale=0.4]{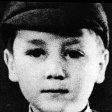}
            \includegraphics[scale=0.4]{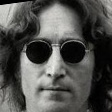}
        }
        \subcaptionbox{\label{p_4}}{
            \includegraphics[scale=0.4]{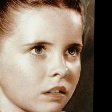}
            \includegraphics[scale=0.4]{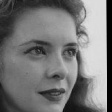}
        }
    }
    \caption{Sample of images across young and old age from AgeDB dataset}
    \label{young_old_pair}
\end{figure}

The motivation of this research is to match the faces of missing persons to their aged counterparts, assisting their recovery and helping families reconnect with long-lost members. The current situation is deeply concerning, with missing person reports reaching critical levels worldwide \cite{msiisngPerson1, msiisngPerson2}. Based on the given problem, the primary focus of this research is age-challenging scenarios and tries to address the changes in the face due to age. Aging brings about several noticeable changes in the face, including hyperpigmentation, sagging skin, changes around the eyes, shifts in facial proportions, wrinkles, and fine lines \cite{face_change1, face_change2}. Additionally, aging affects overall facial size, symmetry, and the texture of the skin, leading to significant visual transformations as shown in Figure \ref{young_old_pair}. This can cause alterations in facial landmarks, including variations in eye width, nasal width or height, and the ratio of facial height to width. Alterations in facial features lead to a substantial shift in the embedding space for the same identity. However, age-variant faces typically maintain a consistent spatial relationship regardless of age and the transformer networks \cite{transformer} are effective at capturing long-range dependencies, which could help in modeling how aging-related changes are distributed across the face. We hypothesize that incorporating a transformer as a loss function is particularly effective in age-challenging scenarios, as the spatial relationships between facial features remain relatively consistent across varying ages.

\vspace{3mm}

 The researchers introduced many architectural changes \cite{inceptionv1, densenet, inceptionFace2, inceptionv2Face, densenetFace} to further investigate CNNs \cite{vggnet, resnet} in machine vision and in face recognition tasks. In these designs, each convolution operation progressively reduces spatial detail in the feature map while enhancing the depth of semantic information, enriching the features with higher-level knowledge at each layer. The reduction in spatial information effectively retains only the most relevant neighborhood details, filtering out redundant or less meaningful data. The semantic knowledge along the channels, also known as abstract-level information, makes the network more robust regarding diffeomorphisms.



For both vision and language models, the contextual relationships between tokens are leveraged by the transformer network \cite{transformer, VIT, detr}, which allowed it to gain popularity and attract the attention of the research community. Our research does not use a transformer as the network backbone but instead explores the transformer as an additional loss module that takes the deep convolution output as an input to feedback the CNN network for better optimization. 

The contribution of this research can be summarized as follows:
\begin{itemize}
    \item The transformer loss is good for addressing age-related challenges due to its capability to capture global and long-range dependencies across feature maps.
    \item Usage of the transformer network as an additional loss instead of the main backbone. We combine the outcome from transformer loss and the standard metric loss to optimize the convolution backbone.
    \item We demonstrate that the proposed combined loss function improves the accuracy on age-diverse validation datasets.
\end{itemize}

\begin{figure}
    \centering
    \setlength{\fboxrule}{0pt}
    \fbox{
        \subcaptionbox{\label{c_1}}{
            \includegraphics[]{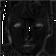}
        }
        \hspace{0.02\textwidth}
        \subcaptionbox{\label{c_2}}{
            \includegraphics[]{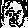} 
        }
        \hspace{0.02\textwidth}
        \subcaptionbox{\label{c_3}}{
            \includegraphics[]{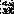} 
        }
        \hspace{0.02\textwidth}
        \subcaptionbox{\label{c_4}}{
            \includegraphics[]{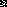} 
        }
    }
    \caption{Convolution output after each bottleneck blocks}
    \label{conv_out}
\end{figure}
\section{Literature Review}



Face recognition, which is a metric learning problem, shifted from softmax loss\cite{center_loss, NoisySoftmax, DeepFace2} into an angular paradigm and proper angular softmax loss function in a variety of forms. The authors \cite{NormFace, LMSoftmax, AMSoftmax, L2EmbedFace} proposed weight and embedding normalization to distribute the embedding into angular space. Incorporating margin in slightly different manners in cosine space \cite{cosface} or in theta space \cite{arcface, sface1} presented better results to the vision community. 

\begin{equation}
\resizebox{0.5\textwidth}{!}{$L_{\textbf{ML}} = - \log \frac{ \exp{ (s \times F ( \cos, \hspace{1px} \theta_{y_i}, \hspace{1px}  m ) } }{ \exp{ (s \times F(\cos, \hspace{1px}  \theta_{y_i} , \hspace{1px}  m )) } + \sum\limits^N_{j=1,j \neq y_i }\exp{(s \times \cos \theta_j})}$}
\label{cosface_fn}
\end{equation}

Other methods also contributed with novel approaches \cite{SymFace, adaface}. By modeling long-range dependencies and contextual relationships, our approach enables the network to better handle variations across age groups and further strengthen identity discrimination.

\section{Method}


In this study, we hypothesize that the final convolution layer can retain contextual information within the neighboring tensor with the help of an appropriate loss function.

We model the association of 3D tensors within the feature map from the final convolution layer beginning with splitting the $H_\text{f} \times W_\text{f} \times D_\text{f}$ feature map into $S_\text{f}$ contextual representation of length $D_\text{f}$. Thus,

\begin{equation}
S_\text{f} = H_\text{f} \cdot W_\text{f}
\label{eq:f_relation}
\end{equation}

At this stage (final convolution), the transformer-based loss is introduced to preserve and model the spatial relationships that CNNs may progressively eliminate as they downsample the final nonlinear feature into a 1-D tensor.

\begin{figure*}
\begin{center}
\fbox{\includegraphics[width=\textwidth]{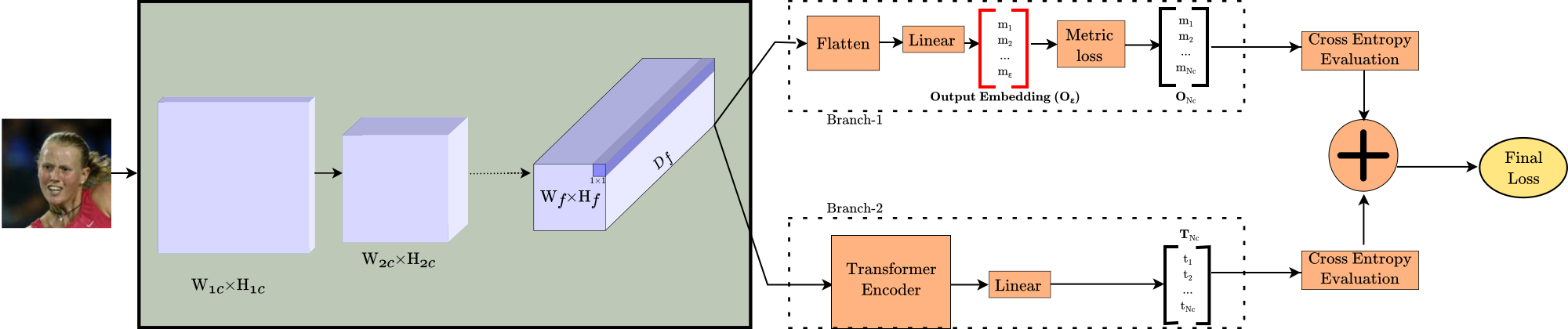}}
\end{center}
   \caption{Transformer-Metric loss architecture}
\label{fig:main_blockdiag}
\end{figure*}

\subsection{Transformer as a Loss}

The proposed architecture passes the input $X$ to the standard CNN backbone, starting with $f$ consecutive CNN layers. The output obtained from the successive CNNs is given as

\begin{equation}
    O_\text{f} = C_{f} 
    \lbrack 
        C_{f - 1} 
        \lbrack 
            \dots 
            C_2 
            \lbrack 
                C_1 
                \lbrack X_i \rbrack 
            \rbrack
        \dots 
        \rbrack
    \rbrack
    \label{eq_v_1}
\end{equation}

where the input $X_i$ is downsampled by the convolution layers into the feature map $O_\text{f}$ of lower spatial dimensions but with a larger depth $D_{\text{f}}$. Before the flattening layer, a split on the network begins from the final convolution layer, namely "branch-1" and "branch-2". The standard branch (branch-1) goes via layer flattening to the final embedding, while the other goes to the proposed transformer block. This is shown in Figure \ref{fig:main_blockdiag}.

The final embedding $O_\varepsilon$ is generated from the standard branch-1, after $O_\text{f}$ is forwarded to flattening and linear transformation $L_{1}$ (refer Eq. \ref{eq_v_2}). Here $O_\varepsilon$ is of size $\varepsilon$, which is the length of the final embedding vector.

\begin{equation}
    O_\varepsilon = L_{1}(O_\text{f})
    \label{eq_v_2}
\end{equation}

In this research, we add a transformer loss (branch-2) from the final convolution layer. We transform the 3-D feature map from feature space $O_f \in \mathbb{R}^{W_f \times H_f \times D_f}$ into contextual vector $V_f \in \mathbb{R}^{S_f \times D_f}$. This sequence of vectors can be treated as sequential patch embedding for the transformer encoder.

We observe a patch size of $1 \times 1$, which is notably smaller than that used in other vision transformers, where the input image is typically divided into larger patches with considerable height and width. Here, we take advantage of CNN to convolve the images into tensors of smaller size ($1 \times 1$) but with a much significant depth $D_\text{f}$. We can accept this sequence of independent contextual vectors of sizes $D_{\text{f}}$ as a condensed representation that retains important context-aware relationships of the input image. Considering the contextual relationship in the latent space of the patch embedding, the depth size $D_{\text{f}}$ plays a significant role in the transformer encoder.

These contextual vectors possess meaningful information that also includes critical relational patterns among each other and feeds this input to the transformer encoder layer, as shown in Eq. \ref{eq_v_3}. Here, we use a standard transformer block composed of six encoder layers. 

\begin{equation}
    T_{S} = \textbf{T-Encoder}( O_{\text{f}} ) \quad T_{S} \in \mathbb{R}^{S_\text{f} \times D_\text{f}}
    \label{eq_v_3}
\end{equation}

\begin{equation}
    T_{S} = \lbrack T_1, T_2, \dots, T_{S_\text{f}} \rbrack
    \label{eq_v_4}
\end{equation}

\begin{equation}
    T_\varepsilon = \frac{1}{S_{\text{f}}} \sum_{i=0}^{S_{\text{f}}}  T_{i}
    \label{eq_v_5}
\end{equation}

\begin{equation}
    T_{N_c} = L_{2} (T_\varepsilon)
    \label{eq_v_6}
\end{equation}

Here, $T_{N_c} \in \mathbb{R}^{S_\text{f} \times D_\text{f}}$. We compute the mean along the sequence length (Eq. \ref{eq_v_4}, \ref{eq_v_5}) and transform into the embedding size of $N_c$ matrix using a linear layer (Eq. \ref{eq_v_6}) without any additional settings of activation or dropout. Here $N_c$ is the number of classes required in the classification task.

The output of Eq. \ref{eq_v_2} ($O_\varepsilon$) generated from branch-1, is forwarded to the standard metric loss (Eq. \ref{eq_v_7}). This is the standard procedure for generating the final linear layer $O_{NC}$ for metric learning problems. Here, $O_{N_c} \in \mathbb{R}^{S_\text{f} \times D_\text{f}}$. We can use any existing metric loss, for instance, CosFace loss \cite{cosface}, ArcFace loss \cite{arcface}.   

\begin{equation}
    O_{N_c} = L_{\textbf{ML}} (O_\varepsilon)
    \label{eq_v_7}
\end{equation}

\subsection{Training Loss}

We can combine both losses as a weighted sum, controlled by a loss balancing factor $\alpha$, to evaluate the final loss $\textbf{L}_{F}$ as shown below

\begin{equation}
    \textbf{L}_{F} = (1 - \alpha) C_\lambda (O_{N_c}) + \alpha C_\lambda(T_{N_c})
    \label{eq_v_8}
\end{equation}
Here, $C_\lambda$ is the cross entropy function that evaluates loss for the output probability distribution (for the target $N_c$ classes) against their actual labels. The range of $\alpha$ is (0, 1). The two embedding vectors of size $N_c$ are obtained, one from standard metric loss and the other from transformer loss. The final loss is a function of both embedding obtained from Eq. \ref{eq_v_6} and Eq. \ref{eq_v_7} to compute the final loss ($\textbf{L}_{F}$) as shown in Eq. \ref{eq_v_8}.  This means that the outputs from both the loss functions are then passed to softmax loss separately. Finally, we add the loss derived from both methods. Here, we note that the embedding that is being used for the validation cycle comes from standard branch-1, which is part of the standard metric loss. So, transformer loss can not be used independently here. As shown, transformer loss can only update the weights via the final convolution layer, leaving the remaining layers in the network backbone unaffected (below the final convolution layer). 
\section{Experiments}



\subsection{Datasets}
These experiments use two datasets: MS1M-arc face \cite{MS-celeb-1M} and WebFace4M (a subset of WebFace 260M) \cite{WebFace}. CASIA-Webface \cite{casia} is used for various experiments for study purposes only. The target age-challenging datasets for this research are CA-LFW \cite{CA-LFW} and AgeDB \cite{AgeDB}, along with LFW \cite{LFW}. 
\begin{table}[htbp] 
    \centering
    \begin{tabular}{|c|c|}
        \hline
        \textbf{Loss} & \textbf{LFW Accuracy} \\
        \hline
        ArcFace \cite{MobileFaceNet} & 99.18  \\ \hline
        Transformer-ArcFace & 99.38\\ \hline
    \end{tabular}
\vspace{5pt}
\caption{Comparison study of transformed-metric loss on Casia-Webface 112X96 on MobileFaceNet}
\label{ablation_study1} 
\end{table}

\begin{table*}[htbp]
    \centering
    \begin{tabular}{|c|c||c|c|c|}
        \hline
        \multicolumn{2}{|c||}{Configuration} & \multicolumn{1}{c|}{} & \multicolumn{2}{c|}{Age-variant datasets} \\
        \hline
        \textbf{Train Data} & \textbf{Loss} & \textbf{LFW} & \textbf{AgeDB} & \textbf{CA-LFW} \\ \hline

        MS1MV2 & CosFace\cite{adaface} & 99.81 & 98.11 & 95.76 \\
        & Transformer-CosFace & 99.83 & \color{red}\textbf{98.20} & \color{red}\textbf{96.00} \\ \hline
       
        MS1MV2 & ArcFace & 99.83 & 98.28 & 95.45 \\
        & Transformer-ArcFace & 99.83 & \color{red}\textbf{98.31} & \color{red}\textbf{96.16} \\ \hline
       
        WebFace4M & AdaFace\cite{adaface} & 99.80 & 97.90 & 96.05 \\
        & Transformer-Adaface & 99.82 & \color{red}\textbf{98.02} & \color{red}\textbf{96.07} \\ \hline
       
        WebFace4M & CosFace\cite{WebFace} & 99.80 & 97.45 & 95.95 \\
        & Transformer-Cosface & 99.82 & \color{red}\textbf{98.00} & \color{red}\textbf{96.00} \\ \hline

        WebFace4M & ArcFace\cite{ArcRefWeb4M} & 99.83 & 97.95 & 96.00 \\
        & Transformer-Arcface & 99.82 & \color{red}\textbf{98.00} & \color{red}\textbf{96.02} \\ \hline

    \end{tabular}
    \vspace{5pt}
    \caption{Verification performance (\%) on age-diverse datasets using ResNet100 with embedding size 512.}
    \label{benchmarks}
\end{table*}


       
       
       



\begin{figure*}
    \centering
    \setlength{\fboxrule}{0pt}
    \subcaptionbox{\label{std_arcface}}{
        \centering
        \includegraphics[width=0.35\linewidth]{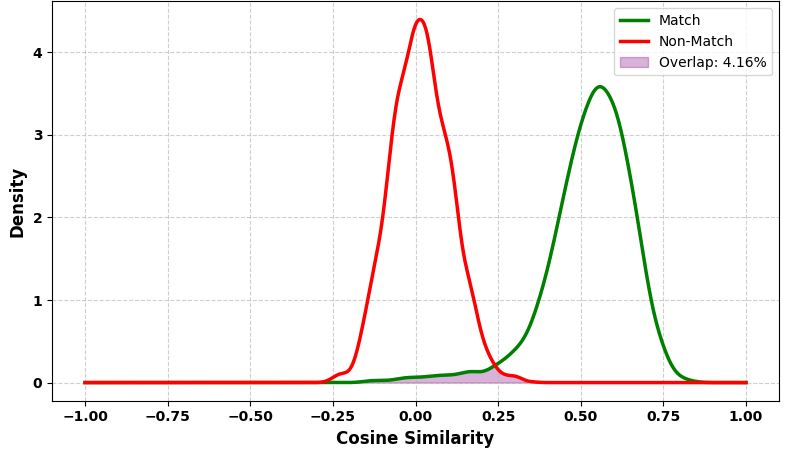}
    }
    \subcaptionbox{ \label{trans_arc}}{
        \centering
        \includegraphics[width=0.35\linewidth]{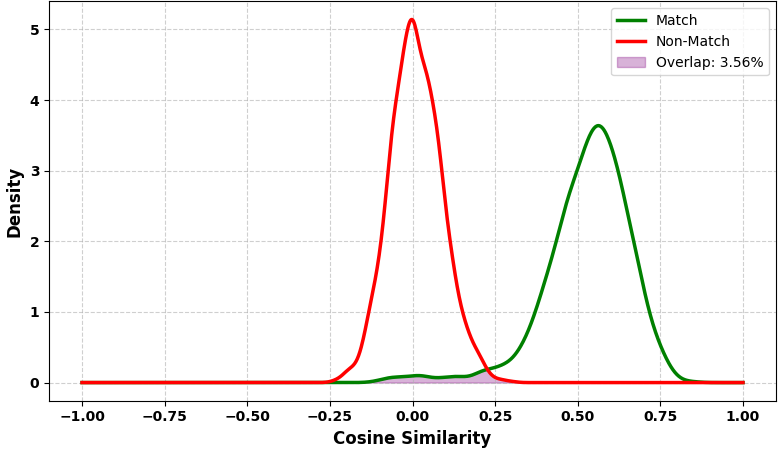}
    }
    \caption{Cosine Similarity Distribution with Match/Non-Match Overlap using ResNet100 on CosFace. Here left graph belongs to base method and right graph belongs to our proposed method}
    \label{fig:plot_analysis}
\end{figure*}

\subsection{Network Settings}

We conduct the experiments on 8 NVIDIA A100 GPUs system. The learning rate is set as 0.1, which gets reduced by a factor of 10 at 10, 18, and 22 epochs. SGD optimizer is used with a momentum value of 0.9, and weight decay is set to 5e-4 for all three neural modules: a) ResNet100, b) Metric loss, and c) Transformer loss. The embedding size of the ResNet network output is set as 512. We set the standard batch size of 512 for the training phase. The face images are normalized by subtracting 127.5 and dividing by 128. Apart from normalization, we flip the image randomly. There are other hyper-parameters that are used by metric loss functions, which we have varied slightly. The value for $s$ is 64 for all losses, but the value of $m$ is set as 0.35 for CosFace, 0.4 for AdaFace, and 0.45 for ArcFace loss.   

The transformer block contains six stacked encoder layers, each takes input of 512 in length. The block has multi-head attention layer (total 8) and feed-forward layer. As the output, the transformer generates an embedding size of 512. 

\subsection{Ablation Study}

In this study, we use the lightweight MobileFaceNet network (0.99M parameter) along with the CASIA-Webface dataset for various experiments. We use ArcFace as metric loss in this experiment, with the goal to optimize the final embedding $O_\varepsilon$ produced by the CNN backbone (Eq. \ref{eq_v_2}). We observe that the transformer-metric loss block gives better performance with an accuracy of 99.38\% compared to the standard metric loss function of 99.18\% as shown in Table \ref{ablation_study1}.



In our experiments, we also find that a very high or very low value of $\alpha$ results in below-average performance on the validation dataset. The value of $\alpha$ in the range of 0.4 to 0.5 shows good accuracy in the LFW dataset.  With these experiments, we can analyze the dominance of the transformer loss function to optimize the CNN network, even if it acts as an additional loss function in metric learning problems. To observe that the order of addition has a significant effect on the training process, when we perform the addition of both embedding vectors before feeding to the softmax loss instead of passing them separately, to the loss function, as shown in Eq. \ref{eq_v_9}, we do not observe any significant progress after certain epochs.

\begin{equation}
    \textbf{L}_{F} = \lambda(O_{N_c} + T_{N_c})
    \label{eq_v_9}
\end{equation}

\subsection{Comparison Results}

The transformer-metric loss functions can be seen performing well on LFW dataset. As per our hypothesis of the addition of a transformer as a loss boosts the performance in age-variant datasets in all the metric loss functions. In Table \ref{benchmarks} observe better results in both of the age-diverse datasets (CALFW and AgeDB) using transformer-metric loss functions. These results prove our hypothesis on age-variant images due to the transformer's ability to model temporal and spatial changes effectively.  Figure \ref{fig:plot_analysis} illustrates the cosine similarity distribution for both the base network and our proposed method, using ResNet100. We observe that our proposed solution significantly reduces the overlap between the match and non-match distributions — often referred to as the "confusion area." Specifically, the overlap in our method occurs only 3.56\% of the time, compared to 4.16\% in the base method. This reduction suggests that our approach achieves better separation between genuine and impostor pairs, which indicates improved discriminative capability and potential for more robust recognition performance.






\section{Discussion}\label{section-discussion}

We compute the inter-class and intra-class variance between the standard metric loss and transformer-metric loss, using the CASIA-Webface dataset for the variance analysis. Table \ref{tab-intra-inter} shows the intra-class and inter-class variance of the distributions of embeddings obtained from both methods; the proposed Transformer method used with ArcFace loss, and the standard ArcFace loss.
The ratio between the inter-class and intra-class scores shows that the proposed method has lower intra-class scores but with higher inter-class variance in comparison, but the Transformer-ArcFace loss edges out with a slightly better performance against the ArcFace loss with the ratio between inter-class and intra-class values. This indicates that the proposed method draws better decision boundaries and marginally better separation relative to standard metric loss methods. 

\begin{table}
    \centering
    \begin{tabular}{|c|c|c|c|}
        \hline
        \textbf{Loss}& $\textbf{Intra}$ & $\textbf{Inter}$ & \textbf{Inter}/\textbf{Intra} \\ & \textbf{class} & \textbf{class} & \textbf{Ratio}\\
        \hline 
        Standard & 5.39 & 5.07 &  0.94\\
        ArcFace Loss & & & \\
        \hline
        Transformer & 4.16  & 3.97 & \textbf{0.95} \\
        ArcFace Loss &  &  & \\
        \hline
    \end{tabular}
    \caption{Comparison of inter-class and intra-class variances on CASIA-Webface with MobileFaceNet}
    \label{tab-intra-inter}
\end{table}

\section{Conclusion}\label{section-conclusion}

Overall, our study lays the groundwork for a new perspective in feature learning: enriching standard discriminative objectives with context-aware transformer guidance, empowering future face recognition systems to achieve higher robustness and fairness across diverse scenarios. Additionally, exploring dynamic or adaptive transformer losses that adjust to sample difficulty or dataset characteristics could make the training even more efficient. The results highlight the transformer's capacity to model long-range dependencies within facial feature maps, which traditional CNN-only pipelines often overlook. Looking ahead, this framework opens several promising research directions.
{
    \small
    \bibliographystyle{IEEEtran}

\begin{thebibliography}{10}
\providecommand{\url}[1]{#1}
\csname url@samestyle\endcsname
\providecommand{\newblock}{\relax}
\providecommand{\bibinfo}[2]{#2}
\providecommand{\BIBentrySTDinterwordspacing}{\spaceskip=0pt\relax}
\providecommand{\BIBentryALTinterwordstretchfactor}{4}
\providecommand{\BIBentryALTinterwordspacing}{\spaceskip=\fontdimen2\font plus
\BIBentryALTinterwordstretchfactor\fontdimen3\font minus \fontdimen4\font\relax}
\providecommand{\BIBforeignlanguage}[2]{{%
\expandafter\ifx\csname l@#1\endcsname\relax
\typeout{** WARNING: IEEEtran.bst: No hyphenation pattern has been}%
\typeout{** loaded for the language `#1'. Using the pattern for}%
\typeout{** the default language instead.}%
\else
\language=\csname l@#1\endcsname
\fi
#2}}
\providecommand{\BIBdecl}{\relax}
\BIBdecl

\bibitem{msiisngPerson1}
\BIBentryALTinterwordspacing
A.~Mewett and S.~D.~M. Thomas, ``Missing children, adolescents and young adults: the relationship between age first missing, subsequent missing person reports and other police-related contacts over a 10-year period,'' \emph{Police Practice and Research}, vol.~26, no.~2, pp. 193--206, 2025. [Online]. Available: \url{https://doi.org/10.1080/15614263.2024.2411712}
\BIBentrySTDinterwordspacing

\bibitem{msiisngPerson2}
\BIBentryALTinterwordspacing
E.~Halford, ``Classifying missing persons cases: an analysis of police risk assessments using multi-dimensional scaling,'' \emph{Police Practice and Research}, vol.~25, no.~5, pp. 612--639, 2024. [Online]. Available: \url{https://doi.org/10.1080/15614263.2024.2330623}
\BIBentrySTDinterwordspacing

\bibitem{face_change1}
\BIBentryALTinterwordspacing
S.~R. Coleman and R.~Grover, ``The anatomy of the aging face: Volume loss and changes in 3-dimensional topography,'' \emph{Aesthetic Surgery Journal}, vol.~26, pp. S4--S9, 01 2006. [Online]. Available: \url{https://doi.org/10.1016/j.asj.2005.09.012}
\BIBentrySTDinterwordspacing

\bibitem{face_change2}
\BIBentryALTinterwordspacing
V.~Ilankovan, ``Anatomy of ageing face,'' \emph{British Journal of Oral and Maxillofacial Surgery}, vol.~52, no.~3, pp. 195--202, 2014. [Online]. Available: \url{https://www.sciencedirect.com/science/article/pii/S0266435613004993}
\BIBentrySTDinterwordspacing

\bibitem{transformer}
A.~Vaswani, N.~Shazeer, N.~Parmar, J.~Uszkoreit, L.~Jones, A.~N. Gomez, L.~Kaiser, and I.~Polosukhin, ``Attention is all you need,'' in \emph{Proceedings of the 31st International Conference on Neural Information Processing Systems}, ser. NIPS'17.\hskip 1em plus 0.5em minus 0.4em\relax Red Hook, NY, USA: Curran Associates Inc., 2017, p. 6000–6010.

\bibitem{inceptionv1}
C.~Szegedy, W.~Liu, Y.~Jia, P.~Sermanet, S.~Reed, D.~Anguelov, D.~Erhan, V.~Vanhoucke, and A.~Rabinovich, ``Going deeper with convolutions,'' in \emph{2015 IEEE Conference on Computer Vision and Pattern Recognition (CVPR)}, 2015, pp. 1--9.

\bibitem{densenet}
G.~Huang, Z.~Liu, L.~Van Der~Maaten, and K.~Q. Weinberger, ``Densely connected convolutional networks,'' in \emph{2017 IEEE Conference on Computer Vision and Pattern Recognition (CVPR)}, 2017, pp. 2261--2269.

\bibitem{inceptionFace2}
S.~Baixo, T.~Ribeiro, G.~Lopes, and A.~F. Ribeiro, ``3d face recognition using inception networks for service robots,'' in \emph{2022 IEEE International Conference on Autonomous Robot Systems and Competitions (ICARSC)}, 2022, pp. 47--52.

\bibitem{inceptionv2Face}
X.~Wan, F.~Ren, and D.~Yong, ``Using inception-resnet v2 for face-based age recognition in scenic spots,'' in \emph{2019 IEEE 6th International Conference on Cloud Computing and Intelligence Systems (CCIS)}, 2019, pp. 159--163.

\bibitem{densenetFace}
A.~Nandy, ``A densenet based robust face detection framework,'' in \emph{2019 IEEE/CVF International Conference on Computer Vision Workshop (ICCVW)}, 2019, pp. 1840--1847.

\bibitem{vggnet}
S.~Liu and W.~Deng, ``Very deep convolutional neural network based image classification using small training sample size,'' in \emph{2015 3rd IAPR Asian Conference on Pattern Recognition (ACPR)}, 2015, pp. 730--734.

\bibitem{resnet}
K.~He, X.~Zhang, S.~Ren, and J.~Sun, ``Deep residual learning for image recognition,'' in \emph{2016 IEEE Conference on Computer Vision and Pattern Recognition (CVPR)}, 2016, pp. 770--778.

\bibitem{VIT}
\BIBentryALTinterwordspacing
Z.~Fu, ``{Vision Transformer: Vit and its Derivatives},'' \emph{arXiv (Cornell University)}, 5 2022. [Online]. Available: \url{https://arxiv.org/abs/2205.11239}
\BIBentrySTDinterwordspacing

\bibitem{detr}
N.~Carion, F.~Massa, G.~Synnaeve, N.~Usunier, A.~Kirillov, and S.~Zagoruyko, ``End-to-end object detection with transformers,'' in \emph{Computer Vision -- ECCV 2020}, A.~Vedaldi, H.~Bischof, T.~Brox, and J.-M. Frahm, Eds.\hskip 1em plus 0.5em minus 0.4em\relax Cham: Springer International Publishing, 2020, pp. 213--229.

\bibitem{center_loss}
Y.~Wen, K.~Zhang, Z.~Li, and Y.~Qiao, ``A discriminative feature learning approach for deep face recognition,'' in \emph{Computer Vision -- ECCV 2016}, B.~Leibe, J.~Matas, N.~Sebe, and M.~Welling, Eds.\hskip 1em plus 0.5em minus 0.4em\relax Cham: Springer International Publishing, 2016, pp. 499--515.

\bibitem{NoisySoftmax}
B.~Chen, W.~Deng, and J.~Du, ``Noisy softmax: Improving the generalization ability of dcnn via postponing the early softmax saturation,'' in \emph{2017 IEEE Conference on Computer Vision and Pattern Recognition (CVPR)}, 2017, pp. 4021--4030.

\bibitem{DeepFace2}
\BIBentryALTinterwordspacing
O.~M. Parkhi, A.~Vedaldi, and A.~Zisserman, ``Deep face recognition,'' in \emph{Proceedings of the British Machine Vision Conference (BMVC)}, M.~W.~J. Xianghua~Xie and G.~K.~L. Tam, Eds.\hskip 1em plus 0.5em minus 0.4em\relax BMVA Press, September 2015, pp. 41.1--41.12. [Online]. Available: \url{https://dx.doi.org/10.5244/C.29.41}
\BIBentrySTDinterwordspacing

\bibitem{NormFace}
\BIBentryALTinterwordspacing
F.~Wang, X.~Xiang, J.~Cheng, and A.~L. Yuille, ``Normface: L2 hypersphere embedding for face verification,'' \emph{Proceedings of the 25th ACM international conference on Multimedia}, 2017. [Online]. Available: \url{https://api.semanticscholar.org/CorpusID:7680631}
\BIBentrySTDinterwordspacing

\bibitem{LMSoftmax}
\BIBentryALTinterwordspacing
W.~Liu, Y.~Wen, Z.~Yu, and M.~Yang, ``Large-margin softmax loss for convolutional neural networks,'' in \emph{Proceedings of the 33nd International Conference on Machine Learning, {ICML} 2016, New York City, NY, USA, June 19-24, 2016}, ser. {JMLR} Workshop and Conference Proceedings, M.~Balcan and K.~Q. Weinberger, Eds., vol.~48.\hskip 1em plus 0.5em minus 0.4em\relax JMLR.org, 2016, pp. 507--516. [Online]. Available: \url{http://proceedings.mlr.press/v48/liud16.html}
\BIBentrySTDinterwordspacing

\bibitem{AMSoftmax}
F.~Wang, J.~Cheng, W.~Liu, and H.~Liu, ``Additive margin softmax for face verification,'' \emph{IEEE Signal Processing Letters}, vol.~25, no.~7, pp. 926--930, 2018.

\bibitem{L2EmbedFace}
\BIBentryALTinterwordspacing
R.~Ranjan, C.~D. Castillo, and R.~Chellappa, ``L2-constrained softmax loss for discriminative face verification,'' 2017. [Online]. Available: \url{https://arxiv.org/abs/1703.09507}
\BIBentrySTDinterwordspacing

\bibitem{cosface}
H.~Wang, Y.~Wang, Z.~Zhou, X.~Ji, D.~Gong, J.~Zhou, Z.~Li, and W.~Liu, ``Cosface: Large margin cosine loss for deep face recognition,'' in \emph{2018 IEEE/CVF Conference on Computer Vision and Pattern Recognition}, 2018, pp. 5265--5274.

\bibitem{arcface}
J.~Deng, J.~Guo, N.~Xue, and S.~Zafeiriou, ``Arcface: Additive angular margin loss for deep face recognition,'' in \emph{2019 IEEE/CVF Conference on Computer Vision and Pattern Recognition (CVPR)}, 2019, pp. 4685--4694.

\bibitem{sface1}
W.~Liu, Y.~Wen, Z.~Yu, M.~Li, B.~Raj, and L.~Song, ``Sphereface: Deep hypersphere embedding for face recognition,'' in \emph{2017 IEEE Conference on Computer Vision and Pattern Recognition (CVPR)}, 2017, pp. 6738--6746.

\bibitem{SymFace}
\BIBentryALTinterwordspacing
P.~Prakash, K.~R. Jerripothula, A.~J. Sam, P.~K. Singh, and S.~Umamaheswaran, ``Symface: Additional facial symmetry loss for deep face recognition,'' 2024. [Online]. Available: \url{https://arxiv.org/abs/2409.11816}
\BIBentrySTDinterwordspacing

\bibitem{adaface}
M.~Kim, A.~K. Jain, and X.~Liu, ``Adaface: Quality adaptive margin for face recognition,'' in \emph{2022 IEEE/CVF Conference on Computer Vision and Pattern Recognition (CVPR)}, 2022, pp. 18\,729--18\,738.

\bibitem{MS-celeb-1M}
Y.~Guo, L.~Zhang, Y.~Hu, X.~He, and J.~Gao, ``Ms-celeb-1m: A dataset and benchmark for large-scale face recognition,'' in \emph{Computer Vision -- ECCV 2016}, B.~Leibe, J.~Matas, N.~Sebe, and M.~Welling, Eds.\hskip 1em plus 0.5em minus 0.4em\relax Cham: Springer International Publishing, 2016, pp. 87--102.

\bibitem{WebFace}
\BIBentryALTinterwordspacing
Z.~Zhu, G.~Huang, J.~Deng, Y.~Ye, J.~Huang, X.~Chen, J.~Zhu, T.~Yang, J.~Lu, D.~Du, and J.~Zhou, ``Webface260m: A benchmark unveiling the power of million-scale deep face recognition,'' in \emph{2021 IEEE/CVF Conference on Computer Vision and Pattern Recognition (CVPR)}.\hskip 1em plus 0.5em minus 0.4em\relax Los Alamitos, CA, USA: IEEE Computer Society, jun 2021, pp. 10\,487--10\,497. [Online]. Available: \url{https://doi.ieeecomputersociety.org/10.1109/CVPR46437.2021.01035}
\BIBentrySTDinterwordspacing

\bibitem{casia}
\BIBentryALTinterwordspacing
D.~Yi, Z.~Lei, S.~Liao, and S.~Z. Li, ``Learning face representation from scratch,'' \emph{CoRR}, vol. abs/1411.7923, 2014. [Online]. Available: \url{http://arxiv.org/abs/1411.7923}
\BIBentrySTDinterwordspacing

\bibitem{CA-LFW}
\BIBentryALTinterwordspacing
T.~Zheng, W.~Deng, and J.~Hu, ``Cross-age {LFW:} {A} database for studying cross-age face recognition in unconstrained environments,'' \emph{CoRR}, vol. abs/1708.08197, 2017. [Online]. Available: \url{http://arxiv.org/abs/1708.08197}
\BIBentrySTDinterwordspacing

\bibitem{AgeDB}
S.~Moschoglou, A.~Papaioannou, C.~Sagonas, J.~Deng, I.~Kotsia, and S.~Zafeiriou, ``Agedb: The first manually collected, in-the-wild age database,'' in \emph{AgeDB: The First Manually Collected, In-the-Wild Age Database}, 07 2017, pp. 1997--2005.

\bibitem{LFW}
G.~B. Huang, M.~Ramesh, T.~Berg, and E.~Learned-Miller, ``Labeled faces in the wild: A database for studying face recognition in unconstrained environments,'' University of Massachusetts, Amherst, Tech. Rep. 07-49, October 2007.

\bibitem{MobileFaceNet}
\BIBentryALTinterwordspacing
J.~Xiao, G.~Jiang, and H.~Liu, ``A lightweight face recognition model based on mobilefacenet for limited computation environment,'' \emph{EAI Endorsed Transactions on Internet of Things}, vol.~7, no.~27, p. 1–9, Feb. 2022. [Online]. Available: \url{https://publications.eai.eu/index.php/IoT/article/view/293}
\BIBentrySTDinterwordspacing

\bibitem{ArcRefWeb4M}
M.~S. Ebrahimi~Saadabadi, S.~Rahimi~Malakshan, A.~Zafari, M.~Mostofa, and N.~M. Nasrabadi, ``A quality aware sample-to-sample comparison for face recognition,'' in \emph{2023 IEEE/CVF Winter Conference on Applications of Computer Vision (WACV)}, 2023, pp. 6118--6127.

\end{thebibliography}

}


\end{document}